\documentclass[conference]{IEEEtran}
\IEEEoverridecommandlockouts
\pdfoutput=1
\usepackage{cite}
\usepackage{amsmath,amssymb,amsfonts}
\usepackage[caption=false, font=footnotesize]{subfig}
\usepackage{algorithmic}
\usepackage{graphicx}
\usepackage{textcomp}
\usepackage{xcolor}
\usepackage{tikz}
\usepackage{booktabs}
\usepackage{textcomp}
\usepackage{multirow}

\usepackage{algorithm}

\usepackage[frozencache=true,cachedir=./minted-output]{minted}

\usetikzlibrary{calc}

\def\BibTeX{{\rm B\kern-.05em{\sc i\kern-.025em b}\kern-.08em
    T\kern-.1667em\lower.7ex\hbox{E}\kern-.125emX}}

\setmintedinline{fontsize=\footnotesize,breaklines, breakafter=_}

\begin{document}

\title{PARTIME: Scalable and Parallel Processing Over Time with Deep Neural Networks
\thanks{This work was partially supported by TAILOR and by HumanE-AI-Net, projects funded by EU Horizon 2020 research and innovation programme under GA No 952215 and No 952026, respectively, by the PRIN 2017 project RexLearn (Reliable and Explainable Adversarial Machine Learning), funded by the Italian Ministry of Education, University and Research (grant no. 2017TWNMH2), by the French government, through the 3IA C\^{o}te d’Azur, Investment in the Future, project managed by the National Research Agency (ANR) with the reference number ANR-19-P3IA-0002, and by NVIDIA AI Technology Center, EMEA, with its support and access to compute resources. We acknowledge the CINECA award under the ISCRA initiative, for the availability of high performance computing resources and support. We thank Giuseppe Fiameni for his useful suggestions.}
}

\author{
    \IEEEauthorblockN{
        Enrico Meloni\IEEEauthorrefmark{1}\IEEEauthorrefmark{2}, 
        Lapo Faggi\IEEEauthorrefmark{1}\IEEEauthorrefmark{2}, 
        Simone Marullo\IEEEauthorrefmark{1}\IEEEauthorrefmark{2}, 
        Alessandro Betti\IEEEauthorrefmark{3}, \\
        Matteo Tiezzi\IEEEauthorrefmark{1},
        Marco Gori\IEEEauthorrefmark{1}\IEEEauthorrefmark{3},
        and Stefano Melacci\IEEEauthorrefmark{1}
    }
    meloni@diism.unisi.it, \{lapo.faggi, simone.marullo\}@unifi.it, alessandro.betti@inria.fr, \\
    mtiezzi@diism.unisi.it, marco.gori@unisi.it, mela@diism.unisi.it
    \IEEEauthorblockA{
        \IEEEauthorrefmark{1}University of Siena, Italy
    }
    \IEEEauthorblockA{
        \IEEEauthorrefmark{2}University of Florence, Italy
    }
    \IEEEauthorblockA{
        \IEEEauthorrefmark{3}Université C\^{o}te d’Azur, France
    }
}

\newcommand{\EM}[1]{{\color{red} Enrico: #1}}

\maketitle

\begin{abstract}
    In this paper, we present PARTIME, a software library written in Python and based on PyTorch, designed specifically to speed up neural networks whenever data is continuously streamed over time, for both learning and inference.
    
    Existing libraries are designed to exploit data-level parallelism, assuming that samples are batched, a condition that is not naturally met in applications that are based on streamed data. Differently, PARTIME starts processing each data sample at the time in which it becomes available from the stream. PARTIME wraps the code that implements a feed-forward multi-layer network and it distributes the layer-wise processing among multiple devices, such as Graphics Processing Units (GPUs). Thanks to its pipeline-based computational scheme, PARTIME allows the devices to perform computations in parallel. At inference time this results in scaling capabilities that are theoretically linear with respect to the number of devices. During the learning stage, PARTIME can leverage the  non-i.i.d. nature of the streamed data with samples that are smoothly evolving over time for efficient gradient computations.
    Experiments are performed in order to empirically compare PARTIME with classic non-parallel neural computations in online learning, distributing operations on up to 8 NVIDIA GPUs, showing significant speedups that are almost linear in the number of devices, mitigating the impact of the data transfer overhead.
    \end{abstract}
    
    
    \section{Introduction}
    \label{sec:intro}
    
    In the last few years, the Machine Learning community strongly increased its attention towards those learning problems that are framed as \textit{continual} or \textit{lifelong}~\cite{surveycontinual}. Even if there exists a large number of recent approaches in such a research direction, this learning setting is still extremely challenging. Real-world applications that are well-suited for continual learning are those that have access to a continuous stream of data, where an artificial agent is not only expected to use the data to make predictions, but also to improve itself and to adapt to changes in the environment, i.e. videos, stream of texts, data from a network of sensors, etc. \cite{BETTI2020275,8943154}. In the case of neural networks, the most extreme and challenging context is one in which a simple online update of model parameters is applied at each time instant, given the information from the last received sample \cite{cal}. 
    
    
    Despite the importance of having access to powerful computational resources for continual learning-based applications, current algorithmic solutions have not been paired with the development of software libraries designed to speed-up learning and inference.
    As a matter of fact, storing and processing portions of the streamed data in a batch-like fashion is the most common way to approach the problem, reusing classic non-continual learning tools. However, the artificial nature of this approach is striking. Motivated by the intuitions behind existing libraries for batched data \cite{gpipe} and by approaches that rethink the neural network computational scheme making it local in time and along the network architecture \cite{lp,tiezzideep,plausible,devmarullo}, we propose a different approach to \textit{pipeline parallelism} specifically built for data sequentially streamed \textit{over time}, where multiple devices work in parallel with the purpose of speeding-up computations. Considering $D$ independent devices, such as $D$ Graphics Processing Units (GPUs), the computational time of a feed-forward deep network empowered by our approach theoretically reduces by a factor $1/D$. 
    We experimentally show that the existing overheads due to data transfer among different devices are constant with respect to $D$ in certain hardware configurations. On the reverse side, 
    the higher throughput obtained by a pipeline parallelism are
    associated with a delay 
    between the forward wave and the backward wave while they propagate through the network that is proportional to $D$, a feature that is not critical in applications in which data samples are non-i.i.d. and smoothly evolve over time. 
    
    The list of contributions of this paper are the following. (1) We describe and share a software library named PARTIME (PARallel processing over TIME)\footnote{https://github.com/sailab-code/partime}, written in Python and PyTorch, specifically designed for efficient continual online learning in multi-GPU architectures using pipeline parallelism. Existing feed-forward multi-layer neural architecture can be easily embraced by our computational scheme, distributing inference and learning among multiple devices. (2) We leverage CUDA Streams\footnote{https://pytorch.org/docs/stable/generated/torch.cuda.Stream.html} paired with CUDA Graphs\footnote{https://pytorch.org/blog/accelerating-pytorch-with-cuda-graphs/} to enable fast kernel scheduling, keeping a very low-level of abstraction of the parallel routines. We also provide automatic load balancing tools, specifically designed for the considered application context. 
    (3) We experimentally evaluate the wall-clock running times with different numbers of devices, investigating the impact of the delays and of the data transfer overhead and comparing several architectures. In some configurations we get empirical speedups that are close to the theoretical ones.
    (4) We experimentally evaluate the quality of learning with different pipeline configurations in continual learning-inspired settings.
    
    This paper is organized as follows. Section~\ref{sec:related} describes the relationships of our work with existing literature and software. Section~\ref{sec:local} formally describes how computations are distributed over time, while Section~\ref{sec:organization} is focused on their implementation and usage details. Section~\ref{sec:experiments} shows experimental results and we draw the conclusions in Section~\ref{sec:conclusions} with a discussion on limits and future extensions.

    \section{Related Work}
    \label{sec:related}
    
    
    The intuition of exploiting data- and/or model-level parallelism to speedup and scale training of deep networks \cite{multigputrainingcnnold,alexweirdtrick} has been largely followed in the last decade \cite{pytorchdistributeddataparalleltraining,deepspeedmodelanddataparallelism}, also focusing on specific tasks or classes of neural nets \cite{megatronlm}. 
    The main motivation that inspired these approaches is the large memory requirements of recent neural models. Naively splitting the models into several components distributed across different devices hinders the overall resource utilization, with devices that are left idle waiting for other devices to complete their job. Pipeline parallelism deals with this under-utilization issue, not only splitting the network across multiple devices (e.g., the first layers in GPU 1, some of the following layers in GPU 2, etc.) but also splitting data into micro-batches and scheduling their forward propagation through the various \textit{stages}, such that each device is left idle the least amount of time~\cite{xpipe}.
    Such computational paradigm yields an improvement on the number of inputs processed per seconds, based on the batch size, the number of stages and the communication overheads between devices. As a matter of fact, existing software libraries are designed for mini-batch mode processing, as they are designed to achieve parallelization over big dataset and stochastic gradient descent.
    A mini-batch is then usually split into several micro-batches, that are identified by the order in which they enter the pipeline. In the following, we review the most popular pipeline parallelism-based libraries: GPipe \cite{gpipe}, PipeDream and Pipedream-2BW \cite{narayanan2019pipedream,narayanan2021pipedream2bw}.
    
\begin{figure}[!ht]
        \centering
        \subfloat[GPipe \cite{gpipe}\label{fig:gpipe_timeline}]{%
            \includegraphics[trim={0 0 0 0.8cm},clip,width=0.8\linewidth]{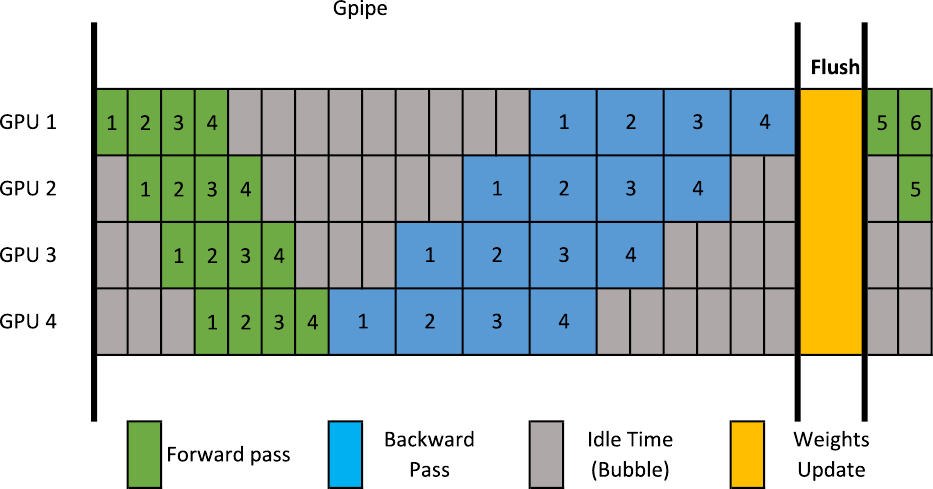}}\\
        \subfloat[PipeDream-2BW \cite{narayanan2021pipedream2bw}\label{fig:2bw_timeline}]{%
            \includegraphics[trim={0 0 0 0.8cm},clip,width=0.8\linewidth]{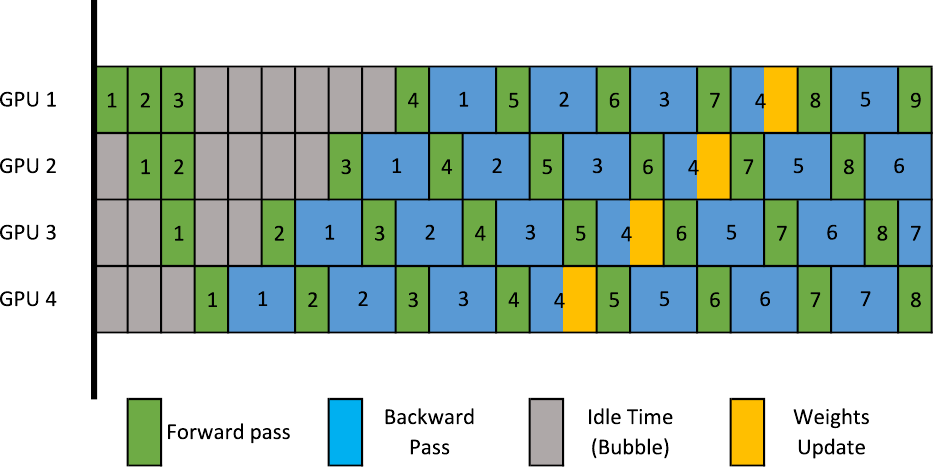}}\\
        \subfloat[PARTIME (this paper)\label{fig:partime_timeline}]{%
            \includegraphics[trim={0 0 0 0.8cm},clip,width=0.8\linewidth]{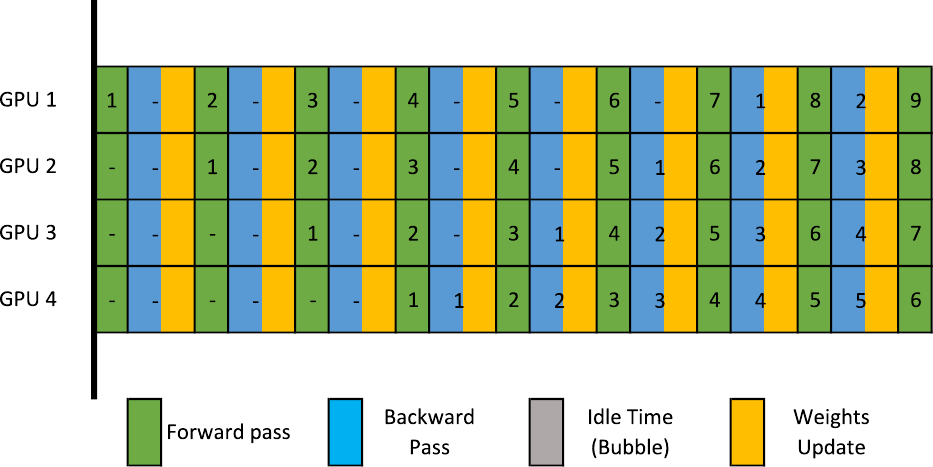}}
        \caption{Processing timelines in different Pipeline implementations. Numbers indicate the ID of an input sample (or micro-batch).}
        \label{fig:pipeline_timeline_comparison}
    \end{figure}

    {GPipe} \cite{gpipe} parallelizes the forward and backward phases across the micro-batches, with the latter being followed by a pipeline flush, where the accumulated gradients are used to update the weights, before moving to the next mini-batch. This induces a significant reduction of the maximum throughput, due to a \textit{bubble} of idle time for most GPUs (see Fig.~\ref{fig:gpipe_timeline} for a visual representation). On the other hand, it allows to faithfully replicate the gradients of a standard sequential processing, without the need to store a copy of the weights. GPipe needs to stash the activation values of each stage for all micro-batches to correctly compute gradients, incurring in a significant memory overhead. 
    {PipeDream} \cite{narayanan2019pipedream} avoids the need of keeping the computations for the forward and backward stages completely separated. Forward and backward processing are interleaved continuously, without leaving any GPU idle. Weights are updated at each step so that, in order to compute the correct gradients for any batch, PipeDream adopts weights stashing in addition to activation stashing. The number of instances of the network weights that must be stored for a pipeline stage $s \in (1, \dots, D)$ is $D - s + 1$, being $D$ the depth of the pipeline.
    PipeDream design allows for a significantly higher throughput with respect to GPipe, but incurs in a heavy memory overhead, which is not suitable for larger models. 
    {PipeDream-2BW} \cite{narayanan2021pipedream2bw} improves the PipeDream design by storing only two versions of the weights in each pipeline stage. A new weight version is generated, by applying the optimizer update rule, every $m \geq D$ micro-batches. However, the novel weight values are not used immediately to avoid inconsistencies in processing the data that already entered the pipeline.
    The new weights are therefore buffered and the oldest version is discarded. PipeDream-2BW introduces a delay between the weights that are used to compute the gradients and the weights that are actually updated. For a network $f$ with weights $W^{(t)}$ for the $t$-th micro-batch, the gradient-based update rule can be expressed simply as $W^{(t+1)} = W^{(t)} - \nu \nabla f(W^{(t-1)})$ (where $\nu$ is the step-size). Thus, PipeDream-2BW tackles the memory overhead, significantly reducing it, and keeps a high speed-up of training times, at the cost of a small approximation of the gradients. See Fig. \ref{fig:2bw_timeline}.
    
    %
    In the case of the proposed {PARTIME}, differently from all the existing approaches (to our best knowledge), we adapt the pipeline parallelism paradigm to the continual online learning scenario, where the input is provided in a sequential manner and there are no offline batches of data. PARTIME does not require mini-batches, and in real-world applications with real-time processing requirements, PARTIME can leverage multiple devices to achieve a processing frame rate closer or equal to the one of the input stream. 
    The computation paradigm is related to that of PipeDream, relaxing the requirement of micro-batches that are now single examples from the input stream. Therefore, forward and backward phases are interleaved continuously without any pipeline flush, allowing a theoretical speed-up of the processing frame-rate of $D$. Differently from PipeDream(-2BW), PARTIME does not stash any activation or weights, further reducing the memory overhead, while keeping a significant speed-up. Differently from GPipe and both the PipeDream approaches, weights are updated at the end of the backward computation of each device, allowing continuous learning of the agent without any hiccup in the framerate, at the cost of an approximation of the gradients, that is not critical when the inputs slowly change over time. We depict in Fig.~\ref{fig:partime_timeline} the PARTIME approach, where the numbers denote the input sample indices, while the sign \textit{-} is used to indicate that the GPU is actually not idle (due to implementation constraints with CUDA Graphs, described in the rest of the paper), even if it using dummy data for the computations.
    
    The theoretical groundings behind the approach implemented in PARTIME can be traced back to early studies on how to distribute the sequential computations of the network layers in a pipeline-like scheme \cite{pipelinednet1993}. More recent activity \cite{plausible,lp} described how such a computational pipeline could be used to train networks where layers are independent modules interconnected by a special class of constraints, and that can learn in parallel. In \cite{plausible} it is shown how different layers can propagate gradients when operating in a parallel manner, exploiting the temporal coherence of streamed data, and that is what we consider in this work. We remark that such a computational structure is also shared by the spatio-temporal local model of \cite{devmarullo}, thus we note that PARTIME could naturally implement also alternatives to Backpropagation, even if this goes beyond the scope of this work.

    \section{Scalable and Parallel Local Computations Over Time}
    \label{sec:local}
    
    
    
    
    
    
    Let us consider a feed-forward neural network composed of $L$ layers. We indicate with $\ell_i(z,w_i)$ the function computed by the $i$-th layer when the input $z$ is provided, while $w_{i}$ is the set of the learnable parameters involved in the layer-internal operations. If $f(x,\omega)$ is the function computed by the network when the input $x$ is provided, being $\omega = \cup_{i=1}^{L} w_i$, we can formally describe $f$ as
    \begin{equation}
        f(x,\omega) = \left(\ell_L(\cdot, w_L) \circ \ell_{L-1}(\cdot, w_{L-1}) \circ \ldots \circ \ell_1(\cdot, w_1) \right)(x).
        \label{eq:f}
    \end{equation}
    Given a scalar loss function $\mathcal{L}$ that depends on the net outputs and other eventually available information $\gamma$ (e.g., supervisions), the Backpropagation algorithm provides a clever way of computing $\nabla_{w_i} \mathcal{L}$, the gradient of $\mathcal{L}$ with respect to some $w_i$, exploiting the composite structure of $f$. Formally, if we indicate with $o^j$ the output of the $j$-th layer,
    \begin{equation}
        \begin{aligned}
            &\nabla_{w_i} \mathcal{L}(f(x,\omega), \gamma) = \frac{\partial \mathcal{L}(o^{L}, \gamma)}{\partial o^{L}} \cdot  \frac{\ell_{L}(o^{L-1}, w_{L})}{\partial w_i} \\
         = & \frac{\partial \mathcal{L}(o^{L}, \gamma)}{\partial o^{L}}  \cdot \frac{\partial \ell_{L}(o^{L-1}, w_{L})}{\partial o^{L-1}} \cdot \frac{\partial \ell_{L-1}(o^{L-2}, w_{L-1})}{\partial o^{L-2}} \cdot\\
         &\qquad \cdots\, \cdot \frac{\partial \ell_{i}(o^{i-1}, w_{i})}{\partial w_i},
        \end{aligned}
    \end{equation}
    where $\cdot$ is the operator involved in the classic chain-rule.\footnote{It is a product when dealing with scalar quantities, otherwise it is the operator that appropriately combines Jacobian matrices or tensors.}
    Let us assume that layers are divided into $D \leq L$ non-overlapping sets also referred to as \textit{stages}, where the $h$-th stage collects consecutive layers with indices in $[a_h,b_h]$, $1 \leq a_h \leq b_h \leq L$, and $a_{h+1} = b_h$. We indicate with $p_h$ the $h$-th stage, involving layer functions $\ell_{a_h}, \ell_{a_h+1}, \ldots, \ell_{b_h}$. 
    The input/output of $p_h$ are $o_{a_h}$ and $o_{b_h}$, respectively.
    Overloading the notation $p_h$ to also refer to the function computed by the $h$-th stage, for simplicity, we can rewrite Eq.~\ref{eq:f} as
    \begin{equation}
        f(x,\omega) = \left(p_D(\cdot, \omega_D) \circ p_{D-1}(\cdot, \omega_{D-1}) \circ \ldots \circ p_1 (\cdot, \omega_1) \right)(x).
        \label{eq:stage}
    \end{equation}
    Let us assume $w_i$ to belong to the weights $\omega_h$ of stage $p_h$, i.e., $b_{h} \leq i \leq a_{h}$. Recalling that $o_{a_h} = o_{b_{h-1}}$, for all valid $h$, the gradient $\nabla_{w_i} \mathcal{L}$ is then
    \begin{equation}
        \begin{aligned}
        \nabla_{w_i}  \mathcal{L}(f(x,\omega), \gamma) &= \frac{\partial \mathcal{L}(o_{b_{D}}, \gamma)}{\partial o_{b_{D}}}  \cdot  \frac{p_{D}(o_{a_{D}}, \omega_{L})}{\partial o_{b_h}} \cdot \frac{p_{h}(o_{a_{h}}, \omega_{h})}{\partial w_i} \\
         &= \frac{\partial \mathcal{L}(o_{b_{D}}, \gamma)}{\partial o_{b_{D}}}  \cdot \frac{\partial p_{D}(o_{a_{D}}, \omega_{D})}{\partial o_{a_{D}}}\\ 
         &\qquad\cdot \frac{\partial p_{D-1}(o_{a_{D-1}}, \omega_{D-1})}{\partial o_{a_{D-1}}} \cdot\\
         &\qquad \qquad\qquad\cdots\, \cdot \frac{\partial p_{h}(o_{a_{h}}, \omega_{h})}{\partial w_i},
        \label{eq:fgrad}
        \end{aligned}
    \end{equation}
    where the last term, $\partial p_{h}(o_{a_{h}}, \omega_{h})/\partial w_i$, can be computed by backpropagating the signal inside the $h$-th stage, 
    \begin{equation}
    \begin{aligned}
           \frac{\partial p_{h}(o_{a_{h}}, \omega_{h})}{\partial w_i} =& \frac{\partial \ell_{b_{h}}(o^{b_{h}-1}, w_{b_{h}})}{\partial o^{b_{h}-1}} \cdot \frac{\partial \ell_{b_{h}-1}(o^{b_{h}-2}, w_{b_{h}-1})}{\partial o^{b_{h}-2}}\cdot\\
           &\quad \cdots\, \cdot \frac{\partial \ell_{i}(o^{i-1}, w_{i})}{\partial w_i}.
        \label{eq:insidestage} 
    \end{aligned}
    \end{equation}
    Whenever data is provided by a generic source stream $\mathcal{S}$, at each time instant $t$ a new sample $x^{(t)}$ is made available, and the following sample is provided after $\Delta_{\mathcal{S}}^{(t)}$ seconds. Without any loss of generality, we will consider the case in which $\Delta_{\mathcal{S}}^{(t)}$ is constant $\forall t$, thus we will just use the notation $\Delta_{\mathcal{S}}$. For example, in the case of a video stream, $\Delta_{\mathcal{S}} = 1/\textsc{fps}$, being \textsc{fps} the frame rate of the video (constant rate). Processing a data sample $x^{(t)}$ requires computing $f(x^{(t)}, \omega)$ and, in a lifelong learning horizon, also $\nabla_{w_i} \mathcal{L}(f(x^{(t)},\omega), \gamma)$, for all the valid $i$'s (plus the weight update operations), that is not an instantaneous processing, especially in very deep networks or when the input size is large. We indicate with $\Delta_{\mathcal{A}}$ such amount of processing time. Whenever $\Delta_{\mathcal{A}} > \Delta_{\mathcal{S}}$, real-time learning is not possible anymore, requiring to skip some frames, since buffering data would just create a constantly increasing queue whose processing is unfeasible. Skipping frames would result in an inherent loss of information that could limit the learning capabilities of the network. As a simple example, consider a network that learns by enforcing motion coherence over consecutive frames \cite{BETTI2020275}. If there is too much distance (in time) between the frames of the pair, there might be poor correlation between them, thus making learning not feasible. Of course, distributing the computations over multiple devices is beneficial for networks that do not fit the memory of a single one, but processing time is still limited by the intrinsic sequential nature of the layer-wise or stage-wise computations, so that a device must wait the previous one to finish its job before starting its own activity.
    
    In this paper we focus on an alternative way of organizing computations over time that is meant for hardware solutions equipped with $D$ computational devices, in particular GPUs, each of them dedicated to the computations of a single stage $p_h$.
    Before going into further details, we remark that what we propose is generic and essentially holds also for other types of devices that can work in parallel or in case of devices that include multiple parallel computational units. While a single GPU actually belongs to the latter category, its parallel computation capabilities are aimed at speeding up lower-level operations (matrix multiplications, convolutions, $\ldots$) that usually exploit most of the GPU resources (e.g., CUDA blocks in NVIDIA cards \cite{cheng2014professional}). As a result, a single GPU is not well-suited for speeding up, for example, multiple stages, even if scheduled for parallel execution.\footnote{In practice, scheduling multiple stages in the same GPU for a potentially parallel execution boils down to the almost serial execution of them, since the lower-level operations within each stage exploit most of the GPU resources. In our experience, there is just a narrow set of cases in which this can lead to non-negligible speed-ups.}
    Let us assume that layer stages are created in a way to have very similar computational times for each stage in the target hardware. Hence, a device that process a single stage in  $\Delta_{\mathcal{P}}$ seconds will require  $\Delta_{\mathcal{A}} = D \Delta_{\mathcal{P}}$ seconds to compute $f(x^{(t)}, \omega)$,  both in the case in which all the stages are sequentially executed on a single device, or if each stage is processed on an independent device, ignoring data transfer overheads. Conversely, a model equipped with pipeline parallelism is ready to process another sample whenever the first GPU has done processing the first stage, that only takes $\Delta_{\mathcal{P}}$, while the vanilla model takes $D\Delta_{\mathcal{P}}$ before being ready to process another sample from the stream. This implies that a pipeline scheme can real-time process the streamed data if $\Delta_{\mathcal{P}} \leq \Delta_{\mathcal{S}}$, instead of $\Delta_{\mathcal{A}} = D\Delta_{\mathcal{P}} \leq \Delta_{\mathcal{S}}$, increasing the throughput by a factor $D$.
    As in classic pipeline parallelism, the $h$-th stage of the pipeline performs the forward and backward phases of the layers of $p_{h}$, using the output that was provided by stage $h-1$ at the previous time step, and the gradients provided by stage $h+1$. In PARTIME, each stage also updates the weight values right after the backward computation has ended (Fig.~\ref{fig:partime_timeline}).
    Thus, it is necessary to include an explicit time dependence in the set of weights characterizing each stage $\omega_i^{(t)}$. Then, the output of stage $i$ at time $t$ will be described by  the temporal index on the values of the weights ($\omega_{\cdot}^{(t)}$), the output of stage $h$ at time $t$ is described as,
    \begin{eqnarray}
    \label{eq:forward}
        o_{b_h}^{(t)} &=& p_h(o_{b_{h-1}}^{(t-\Delta_{\mathcal{P}})},\omega_h^{(t)})\\ 
        \nonumber & \hskip -1.5cm =& \hskip -1.0cm \left(p_h(\cdot,\,\omega_h^{(t)})\circ \cdots \circ p_1(\cdot,\,\omega_1^{(t-(h-1)\Delta_{\mathcal{P}}})\right)(x^{(t-(h-1)\Delta_{\mathcal{P}})})),
    \end{eqnarray}
    According to the above equation, the network output at time $t$ is given by $o_{b_D}^{(t)} = p_{D}(o_{b_{D-1}}^{(t-1)}, \omega_{D}^{(t)})$, and it is equal to
    \begin{equation}
    \begin{aligned}
        &f(x^{(t-(D-1)\Delta_{\mathcal{P}})},\{\omega_D^{(t)},\,\omega_{D-1}^{(t-\Delta_{\mathcal{P}})},\,\dots,\,\omega_{1}^{(t-(D-1)\Delta_{\mathcal{P}})}\})\\
        =&\left(p_D(\cdot, \omega_D^{(t)}) \circ \cdots \circ p_1(\cdot,\,\omega_1^{(t-(D-1)\Delta_{\mathcal{P}})})\right)(x^{(t-(D-1)\Delta_{\mathcal{P}})})\,,
    \end{aligned}
    \label{eq:w}
    \end{equation}
    that is essentially the analogous of \eqref{eq:stage} when making explicit the different time indices attached to the weights involved in processing a sample from the input stream. Of course,  as in every pipeline-based model, the output associated to each sample becomes available with a delay that is $D\Delta_{\mathcal{P}}$, not a crucial issue in case of smoothly evolving streams with relatively large frame rates, as the ones we consider in this paper. 
    
    The backward propagation in PARTIME follows a similar approach to the forward case, with gradients propagating through the pipeline from the loss function down to the first stage. Assuming $w_i$ to belong to the weights $\omega_h$ of stage $p_h$ we generalize \eqref{eq:fgrad} to
    \begin{equation}\medmuskip-3mu
    \label{eq:backprop_with_delay}
    \begin{aligned}
    (\nabla_{w_i} \mathcal{L})^{(t)} = & \left.\frac{\partial \mathcal{L}(o_{b_D}, \gamma^{(t-(D-h)\Delta_{\mathcal{P}})})}{\partial o_{b_D}}\right|_{o_{b_D}^{(t-(D-h)\Delta_{\mathcal{P}})}}\\
    &\cdot\left.\frac{\partial p_D(o_{b_{D-1}}, \omega_D^{(t-(D-h)\Delta_\mathcal{P})})}{\partial o_{b_{D-1}}}\right|_{o_{b_{D-1}}^{(t-(D-h+1)\Delta_\mathcal{P})}}\\
    \cdots\,&\left.\frac{\partial p_{h+1}(o_{b_h}, \omega_{h+1}^{(t-\Delta_\mathcal{P})})}{\partial o_{b_h}}\right|_{o_{b_h}^{(t)}} 
    \cdot\frac{\partial p_{h}(o_{b_{h-1}}^{(t-\Delta_{\mathcal{P}})}, \omega_{h})}{\partial w_{i}}\biggl|_{w_i^{(t)}}\mskip-20mu.
    \end{aligned}
    \end{equation}
    The last term $\partial p_{h}(o_{b_{h-1}}^{(t-\Delta_{\mathcal{P}})}, \omega_{h}))/\partial w_{i}|_{w_{i}^{(t)}}$ is evaluated by means of a backpropagation of the signal within the given stage $h$ according to \eqref{eq:insidestage}, while the product of the previous terms is propagated through the stages. 
    Notice that, focusing on a specific stage $h$, the PARTIME computational structure is characterized by a delay consisting in $2(D-h)$ steps in-between the corresponding forward and backward ``waves'', as can be seen in Fig.~\ref{fig:partime_timeline}, checking each stage/GPU-line and counting the steps between data with the same index---recall that each step consists of forward, backward, update. This leads to an approximation in the evaluation of the gradients since, during such a time interval, the input of the stages changes. 
    Of course, the slower the input stream is varying, the less impacting is the approximation. The delay is zero for $h=D$ since, in this case, the last stage computes the forward, backward (and update) operations related to a given input at the same time. 
    However, it is not only a matter of changing the stage input, since a similar consideration holds for the values of weights, that get updated (thus they change) at each computational step, as we already anticipated in (\ref{eq:w}). 
    Even if this will play a role in the evaluation of the gradients, a small learning rate can mitigate abrupt changes in the values of the weights, making the resulting approximation more appropriate. 
    In the next section we will describe the implementation of the ideas here presented. To this purpose, let us note that we can also write \eqref{eq:backprop_with_delay} as 
    \begin{equation} \label{eq:w_comp}
        (\nabla_{w_i} \mathcal{L})^{(t)} = g_h^{(t)}\cdot\left.\frac{\partial p_{h}(o_{b_{h-1}}^{(t-\Delta_{\mathcal{P}})}, \omega_{h})}{\partial w_{i}}\right|_{w_i^{(t)}}\,,
    \end{equation}
    where we have contextually defined
    \begin{equation}
    \begin{aligned}
        g_h^{(t)} = & \left.\frac{\partial \mathcal{L}(o_{b_D}, \gamma^{(t-(D-h)\Delta_{\mathcal{P}})})}{\partial o_{b_D}}\right|_{o_{b_D}^{(t-(D-h)\Delta_{\mathcal{P}})}}\\
    &\cdot\left.\frac{\partial p_D(o_{b_{D-1}}, \omega_D^{(t-(D-h)\Delta_\mathcal{P})})}{\partial o_{b_{D-1}}}\right|_{o_{b_{D-1}}^{(t-(D-h+1)\Delta_\mathcal{P})}}\\
    \cdots&\cdot\left.\frac{\partial p_{h+1}(o_{b_h}, \omega_{h+1}^{(t-\Delta_\mathcal{P})})}{\partial o_{b_h}}\right|_{o_{b_h}^{(t)}}
        \end{aligned}
    \end{equation}
    that, starting from $ g_D^{(t)} = \partial \mathcal{L}(o_{b_D}, \gamma^{(t)})/\partial o_{b_D}|_{o_{b_D}^{(t)}}$, can be also expressed through to the following recursive relation 
    \begin{equation}
    \label{eq:g_comp}
       g_h^{(t)} =g_{h+1}^{(t-\Delta_\mathcal{P})} \cdot\left.\frac{\partial p_{h+1}(o_{b_h}, \omega_{h+1}^{(t-\Delta_\mathcal{P})})}{\partial o_{b_h}}\right|_{o_{b_h}^{(t)}}\,.
    \end{equation}

    \begin{figure}
        \centering
        \includegraphics[width=0.8\linewidth]{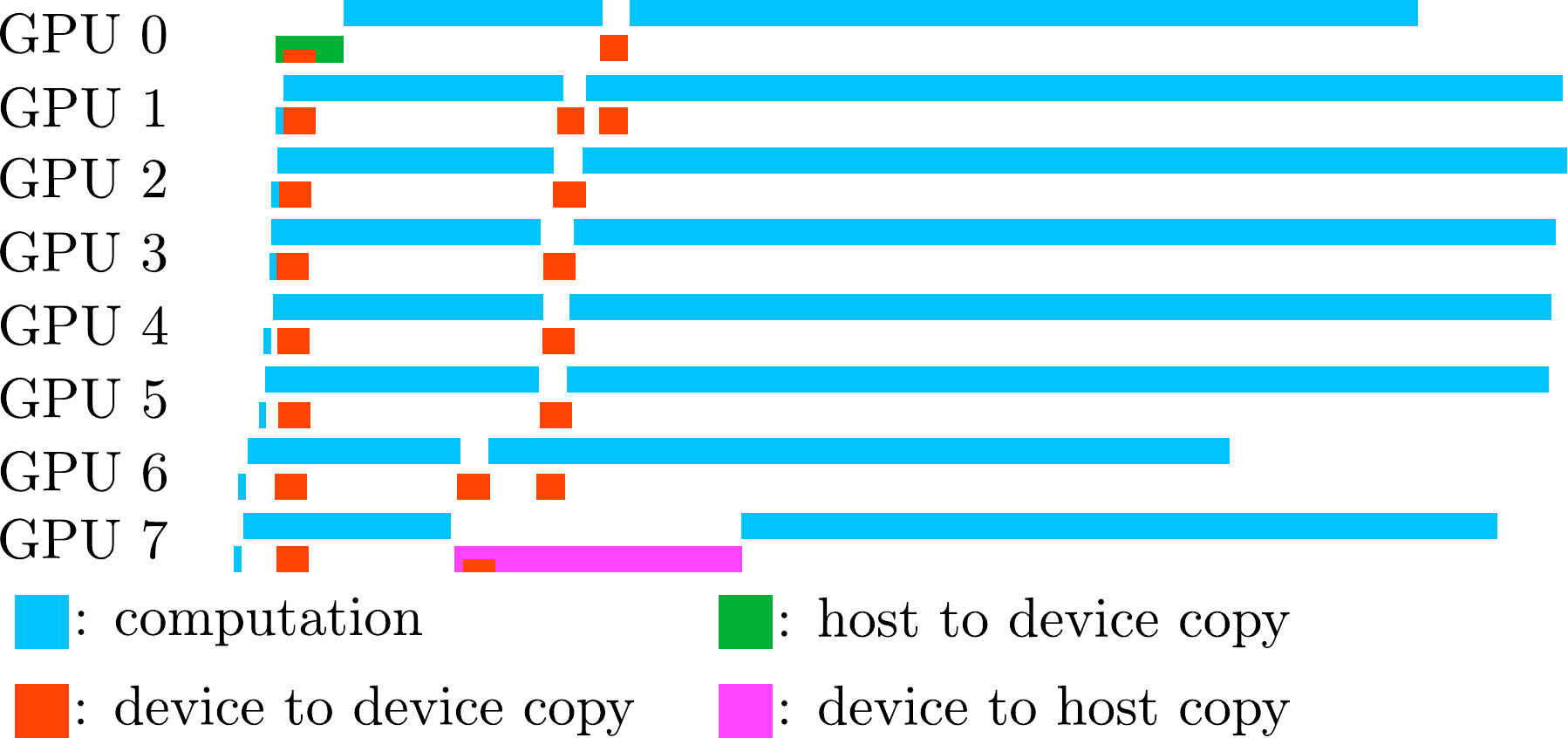}
        \caption{Capture of a single forward/backward step of PARTIME. With the use of CUDA Graphs, every GPU starts processing almost at the same time, reducing communications overhead at its minimum. The result is an almost 8x speed-up. 
        }
        \label{fig:profiling}
    \end{figure}

    \section{PARTIME}\label{sec:organization}
    The PARTIME software library is written in PyTorch\footnote{We exploited PyTorch 1.11 for our implementation.} and it expects the user to provide a classic network of type \texttt{torch.nn.Sequential}, that is automatically converted into a format that will enable pipelined computations, both in the forward and backward phases. Despite the general formulation of the ideas behind PARTIME, the current version of the library leverages  CUDA-based facilities, in particular CUDA Streams and CUDA Graphs, to setup the parallel execution scheme, thus it is designed for NVIDIA GPUs. In Algorithm~\ref{alg:cap} we provide a high-level description of the operations performed by each stage/device.
    
    In order to activate the concurrent execution of independent queues of GPU tasks, PARTIME creates multiple CUDA Streams on each device. Each stream holds a queue of sequential tasks which are executed in-order, while the different streams interleave their tasks in no specified order. Since CUDA Streams are handled in an asynchronous manner (i.e., they do not wait the results of each enqueued task, even if such results are used in the rest of the code), PARTIME relies on specifically placed ``events'' (provided by the CUDA APIs) to ensure proper synchronization of the different streams.
    \begin{algorithm}
    \caption{Operations performed by each device/stage. Each stage expects input data to be stored in the so-called {\it stable} memory area, and it also has the use  of a {\it temporary} one.}\label{alg:cap}
    \small
    \begin{algorithmic}[1]
    \IF{1st stage}
    \STATE Copy $x^{(t)}$ from host memory to the {\it stable} input memory area.
    \ELSE
    \STATE Copy {\it temporary} memory area to {\it stable} input memory area.
    \ENDIF
    \STATE Start forward step, using the  {\it stable} input memory \{Eq.~\eqref{eq:forward}\}.
    \IF{last stage}
    \STATE When 6 completes, copy last stage output to host memory.
    \ELSE
    \STATE Copy gradients from the next stage 
    \COMMENT{$g^{(t-\Delta_\mathcal{P})}_{h+1}$ in Eq. \eqref{eq:g_comp}}
    \STATE When 6 completes, copy output to the next-stage {\it temporary} input memory area (to avoid overwriting {\it stable} data--currently used by the next-stage).
    \ENDIF
    \STATE Compute stage-related gradients \COMMENT{Eq. \eqref{eq:w_comp}}
    \STATE Update weights.
    \end{algorithmic}
    \end{algorithm}
    A major drawback of a vanilla CUDA Stream-based implementation is the communication-overhead introduced when the CPU enqueues a task to the GPU streams. Such overhead quickly becomes negligible when dealing with tasks that run for a long time, but not when processing streams of data with the purpose of splitting computations into fast-processing parallel stages, where (in some configurations) enqueueing could take more time than the actual execution of each stage. 
    PARTIME bypasses this issue exploiting the recently introduced NVIDIA CUDA Graphs, that are able to handle the dependencies among the various GPU tasks in a warmup stage, generating a compact Directed Acyclic Graph (DAG) that summarizes all the operations. This makes all the GPU-task virtually collapse into a single one, with a scheduling overhead that is paid only once for all the wrapped tasks. Since CUDA Graphs might not be supported by older GPUs, PARTIME can easily switch-back to the case in which they are not used.
    
    In Fig.~\ref{fig:profiling} we report the outcome of an operation-capture performed with the NVIDIA NSIGHT System, of a single pipeline step, confirming that all the GPUs are able to work in parallel, distributing the computational burden in an effective manner. 
    Of course, in order to get the most out of the pipeline scheme, the computational time of the different stages should be comparable, to avoid the slower stage to reduce the throughput. PARTIME provides a stage balancing procedure, in which the network layers are automatically partitioned, that is what we used both in Fig.~\ref{fig:profiling} and in the following experiments of this paper. First of all, the copy-times from/to host memory to GPUs are measured, assigning the first/last pipeline stages to the fastest GPUs (the speed depends on the hardware configuration of the GPUs in the host machine). Then, PARTIME evaluates the processing times of all the layers in each GPU, and
     it splits layers into stages trying to make their overall computational times similar, under some constraints,  with the same algorithm exploited in related libraries \cite{kim2020torchgpipe} that, differently from PARTIME, do not consider data-transfer times.
    
    
    In the PARTIME library, a \texttt{Pipeline} object holds an ordered list of \texttt{Stages}.  
    Each of the \texttt{Stages} contains a non-overlapping portion of the original network. 
    The \texttt{Pipeline} is initialized using ($i$) the neural network model to be handled, ($ii$) the number of stages (eventually, how layers should be distributed),
    ($iii$) the list of available GPU devices (for maximum performance, it should be the same as the number of stages, but it can be smaller, making multiple stages execute on the same device), 
    ($iv$) the optimizer settings for weight updates (eventually, a user defined loss functions), and ($v$) a sample tensor whose shape is leveraged to infer the shapes of static memory allocations, required by CUDA Graphs.
    \texttt{Pipeline} objects provide a \texttt{forward} method that processes inputs with the same shape of the previously described sample tensor, triggering the computations described in Algorithm~\ref{alg:cap} over all the \texttt{Stages} (forward, backward, update). 
    In the following we provide a simple code snippet with an example of usage of PARTIME, showing that a few lines of code are needed to setup a pipeline-based execution.
    \begin{minted}[frame=lines,bgcolor=white,fontsize=\scriptsize]{python}
    from partime.pipeline import Pipeline
    from partime.balancing import balance_pipeline_partitions
    
    net = YourSequentialModel()
    balance = balance_pipeline_partitions(net,
        devices, len(devices))
    
    pipeline = Pipeline(
      net,           # Network to be wrapped
      sample_input,  # A tensor with the 
                     #  same shape as the input stream
      balance,       # A list that determines how 
                     #  layers are split, e.g. [8, 10, 12, 11]
      devices,       # List of devices to use
      cuda_graph,    # Flag to enable CUDA Graph
      loss_fn,       # A callable that returns the loss
      sample_target, # A tensor with the 
                     #  same shape as the targets
      optim_settings # Tuple with optimizer class 
                     #  and dict of hyper-parameters
    )
    
    for idx, (inp, target) in enumerate(stream):
      pipeline.forward(inp, target)
      if idx < len(pipeline.stages) - 1:
        continue # The first input still has not reached
                 #  the end of the pipeline
      else:
        outputs = pipeline.outputs_buffer
        loss = pipeline.loss_buffer
        # Print/display/log loss and output
    \end{minted}
    
    
    \section{Experiments}
    \label{sec:experiments}
    
    We performed several experiments to showcase the processing speedup obtained via the PARTIME library, starting with a stream of visual data, composed of $10000$ frames. We evaluated the speed at which the network performs \textit{inference} (forward only), 
    comparing it to the time needed for the same operations with vanilla sequential (i.e., non-pipeline-based) models. 
    Moreover, we also considered the case in which the PARTIME pipeline is used to wrap a neural model that performs continual online {\it learning}, thus also including the backward and update steps. Then, we evaluated the quality of the trained model in a continual online motion estimation experiment and in a classification problem based on a stream of images from the CIFAR-10 dataset, providing the network with data taken from a sliding window of samples, to avoid abrupt changes in input at consecutive time steps.

    \noindent\textbf{A. Pixel-wise Predictions.}
    Our first experimental activity is about a neural architecture composed by $150$ convolutional layers having fixed input/output resolution (i.e.~no pooling or stride $>1$), thus simulating the prediction of pixel-wise features. We assume to deal with input tensors having a squared spatial resolution of $R\times R$ pixels, with $R \in \{256, 1024\}$ and 3 channels. 
    %
    Table~\ref{tab:inferenceA} (top) shows the speedup in {\it inference}, averaged over the considered forward steps, reporting in bold those speedups that are greater than 80\% of the theoretical speedup. 
    We considered different settings, consisting in various combinations of input resolutions ($R \times R$) and number of output features ($F \in \{1, 10, 100\}$)  to better evaluate the impact of the communication overheads with different data sizes. We denoted each setting with the compact notation $R / F$. We also considered varying  numbers of pipeline stages (\textsc{\#Stages} $\in \{2, 4, 8\}$). The obtained results confirm the huge contribution of CUDA Graph in the low-resolution settings, where communications overheads are more impactful than computational times, while advantages of CUDA Graph are less evident with bigger input resolutions. In all the cases PARTIME provides significant improvements over the vanilla sequential network, reaching results closer to the theoretical case. 
    %
    Noticeably, with $R=1024$ the exploitation of PARTIME yields a speedup $\approx \times 7$, close to the maximum theoretical speed-up of $8$. In  Table~\ref{tab:inferenceA} (bottom) we report the {\it learning} case, in which also the backward phase and optimization step are included. Speedups are even greater than the inference case, as backward computations increase the computational cost of each single pipeline stage, reducing the relative impact of data-transfer overhead. 

    \begin{table}[t]
        \centering
        \caption{Pixel-wise predictions. Speed-up achieved by PARTIME considering only inference (top) and both forward and backward learning phases (bottom), varying the input tensor resolution $R$ and the number of output channels $F$ (rows), as well as the number of pipeline stages (columns), and investigating the advantages of CUDA Graph.}
        \begin{tabular}{ccrrrrrr}
            \toprule
           & {} & \multicolumn{3}{l}{Without CUDA Graph} & \multicolumn{3}{l}{With CUDA Graph} \\
           \cmidrule(l){3-5} \cmidrule(l){6-8}
            & \textsc{\#Stages} &                  2 &    4 &    8 &               2 &    4 &    8 \\
            & R / F &                    &      &      &                 &      &      \\
            \midrule
             \parbox[t]{2mm}{\multirow{6}{*}{\rotatebox[origin=c]{90}{Inference Only}}} 
            &256 / 1      &               1.12 & 1.17 & 1.15 &            \textbf{1.89} & \textbf{3.74} & \textbf{6.41} \\
            &256 / 10      &               1.09 & 1.12 & 1.10 &            \textbf{1.68} & \textbf{3.65} & \textbf{6.46} \\
            &256 / 100      &               1.08 & 1.11 & 1.08 &            \textbf{1.72} & \textbf{3.13} & 5.05 \\
            &1024 / 1     &               \textbf{1.82} & \textbf{3.28} & 5.27 &            \textbf{1.86} & \textbf{3.87} & \textbf{7.14} \\
            &1024 / 10     &               \textbf{1.82} & \textbf{3.21} & 5.53 &            \textbf{1.88} & \textbf{3.57} & \textbf{6.65} \\
            &1024 / 100     &               \textbf{1.72} & 2.79 & 4.44 &            \textbf{1.69} & \textbf{2.99} & 4.81 \\
            \midrule
             \parbox[t]{2mm}{\multirow{6}{*}{\rotatebox[origin=c]{90}{Learning}}} 
            &256 / 1      &               1.36 & 1.45 & 1.46 &            \textbf{1.96} & \textbf{3.80} & \textbf{7.16} \\
            &256 / 10      &               1.38 & 1.42 & 1.48 &            \textbf{1.99} & \textbf{4.1}1 & \textbf{7.45} \\
            &256 / 100      &               1.28 & 1.46 & 1.48 &            \textbf{1.99} & \textbf{3.57} & \textbf{6.70} \\
            &1024 / 1     &               \textbf{1.81} & \textbf{3.43} & 5.63 &            \textbf{1.95} & \textbf{3.66} & \textbf{6.61} \\
            &1024 / 10     &               \textbf{1.85} & \textbf{3.63} & \textbf{6.26} &            \textbf{1.95} & \textbf{3.81} & \textbf{7.37} \\
            &1024 / 100     &               \textbf{1.88} & \textbf{3.39} & 5.57 &            \textbf{1.89} & \textbf{3.57} & \textbf{6.35} \\
            \bottomrule
        \end{tabular}
        
        \label{tab:inferenceA}
    \end{table}

    \noindent\textbf{B. Image Classification.}
    The second experimental activity is about neural architectures composed of $150$ convolutional layers interleaved with pooling layers, with the final output pooled to a vector with $F$ elements, the number of output classes (1 vector per image). The main goal of this activity is to evaluate potential pipeline balancing issues due to the different spatial resolutions of the layers (i.e. layers processing inputs with spatial resolution smaller than $R$ have less impact in the overall computation). 
    Table~\ref{tab:inferenceB} (top, same structure of Table \ref{tab:inferenceA}, reporting in bold the cases in which the speedups are greater than half of the theoretical one) shows that CUDA Graph yields the best results even in this experience. The aforementioned layer/stage balancing issue causes an under-utilization of some of the GPUs, that is more evident in the case of $R=1024$, where the maximum speed-up is almost $\times 3$ even with 8 GPUs.
    

    \noindent\textbf{C. Image Classification with Residual Connections.}
    We performed another image classification experience using a ResNet-152 architecture, customizing the final classification head to yield $F$ output classes/features. Skip connections from layer $i$ to layer $j>i$ are propagated through all the stages in-between $i$  and $j$, by means of identity mappings. Therefore, we remark that skip connections introduce further copy operations between GPU devices. Nonetheless, the speed-up achieved by PARTIME are similar to the previous experience 
    Settings with \textsc{\#Stages} $\in \{2, 4\}$ get closer to the theoretical speed-up, since the data that is about the skip connections need to be propagated through less stages.
    

    \begin{table}
    \centering
    \caption{Image Classification without and with residual connections (top and bottom part of the table, respectively). See Table \ref{tab:inferenceA} for a description of the labels.}
    \begin{tabular}{ccrrrrrr}
    \toprule
    &{} & \multicolumn{3}{l}{Without CUDA Graph} & \multicolumn{3}{l}{With CUDA Graph} \\
    \cmidrule(l){3-5} \cmidrule(l){6-8}
    &\textsc{\#Stages} &                  2 &    4 &    8 &               2 &    4 &    8 \\
    &R / F &                    &      &      &                 &      &      \\
    \midrule
     \parbox[t]{2mm}{\multirow{6}{*}{\rotatebox[origin=c]{90}{Classifier}}} 
    &256 / 1      &               1.16 & 1.13 & 1.06 &            0.78 & \textbf{2.59} & \textbf{4.31} \\
    &256 / 10      &               1.11 & 1.04 & 1.00 &            0.99 & \textbf{2.55} & 3.48 \\
    &256 / 100      &               1.10 & 1.06 & 1.00 &            \textbf{1.69} & \textbf{2.67} & \textbf{4.39} \\
    &1024 / 1     &               \textbf{1.63} & 1.67 & 1.18 &            \textbf{1.43} & \textbf{2.61} & 2.89 \\
    &1024 / 10     &               \textbf{1.66} & 1.64 & 1.38 &            0.80 & 1.64 & 2.52 \\
    &1024 / 100     &               \textbf{1.43} & 1.63 & 1.49 &            \textbf{1.20} & 1.24 & 2.79 \\
    \midrule
     \parbox[t]{2mm}{\multirow{6}{*}{\rotatebox[origin=c]{90}{ResNet}}} 
    &256 / 1    &               \textbf{1.10} & 1.07 & 1.04 &            \textbf{1.40} & \textbf{2.94} & 2.11 \\
    &256 / 10   &               \textbf{1.11} & 1.04 & 1.02 &            \textbf{1.77} & \textbf{2.99} & \textbf{4.50} \\
    &256 / 100  &               \textbf{1.11} & 1.06 & 0.98 &            \textbf{1.66} & \textbf{2.93} & \textbf{4.31} \\
    &1024 / 1   &               \textbf{1.24} & 1.23 & 0.99 &            \textbf{1.42} & \textbf{2.34} & 1.50 \\
    &1024 / 10  &               \textbf{1.23} & 1.20 & 1.15 &            \textbf{1.07} & \textbf{2.72} & 3.88 \\
    &1024 / 100 &               \textbf{1.27} & 1.22 & 1.14 &            \textbf{1.69} & \textbf{3.28} & 2.94 \\
    \bottomrule
    \end{tabular}
    
        \label{tab:inferenceB}
    \end{table}

\begin{figure}
        \centering
        \includegraphics[width=0.8\linewidth]{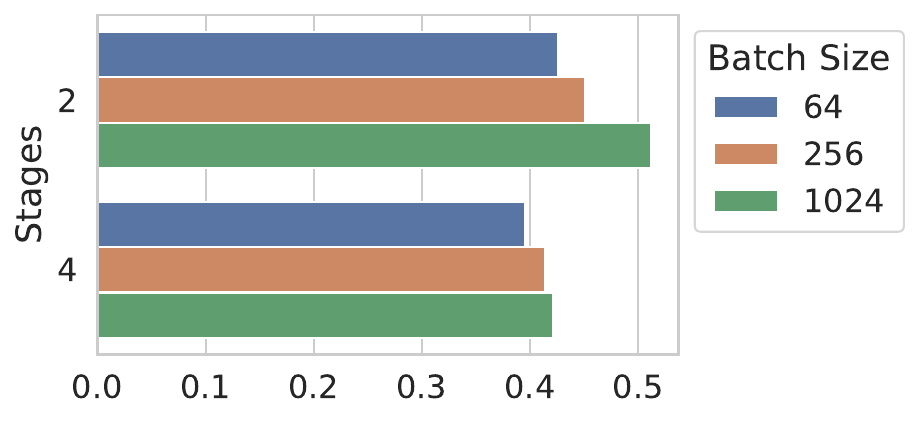}
        \caption{Test accuracy in the continual online image classification described in the main text. For reference, the baseline accuracies for a sequential network trained without pipelining are $0.489$, $0.553$, and $0.639$ for batch sizes of $64$, $256$ and $1024$, respectively.}
        \label{fig:final_acc}
    \end{figure}

    \noindent\textbf{D. Continual Online Image Classification.}
    In our next experience we simulated a continual online learning process where a neural model is trained on the CIFAR-10 dataset, with examples provided in a sequential manner. The training procedure collects inputs using a sliding window to build up a ``replay batch'', which is then fed to the pipeline. Each time a new sample is provided, the oldest input in the replay batch is replaced by the new one, while the rest is kept as is. This ensures that the PARTIME assumption of processing slowly-varying inputs is approximately met. 
    The learning rate is also chosen accordingly to avoid changing the weights too much between each step. 
    We considered a ResNet-50 with 10-classes classification head, trained using the Adam optimizer with learning rate $\mu=0.001$, streaming the whole dataset 5 times, testing different sliding-window/batch sizes in $\{ 64, 256, 1024\}$. 
     Fig.~\ref{fig:final_acc} compares the accuracy on the test data for the different batch sizes. 
    Training a 2-stage pipelined network takes around half of the time of the sequential model. The results show that when increasing the batch size, the pipeline is able to learn better predictors, as we better implement the slowly-changing input condition.

\begin{figure}
        \centering
        \includegraphics[width=0.7\linewidth]{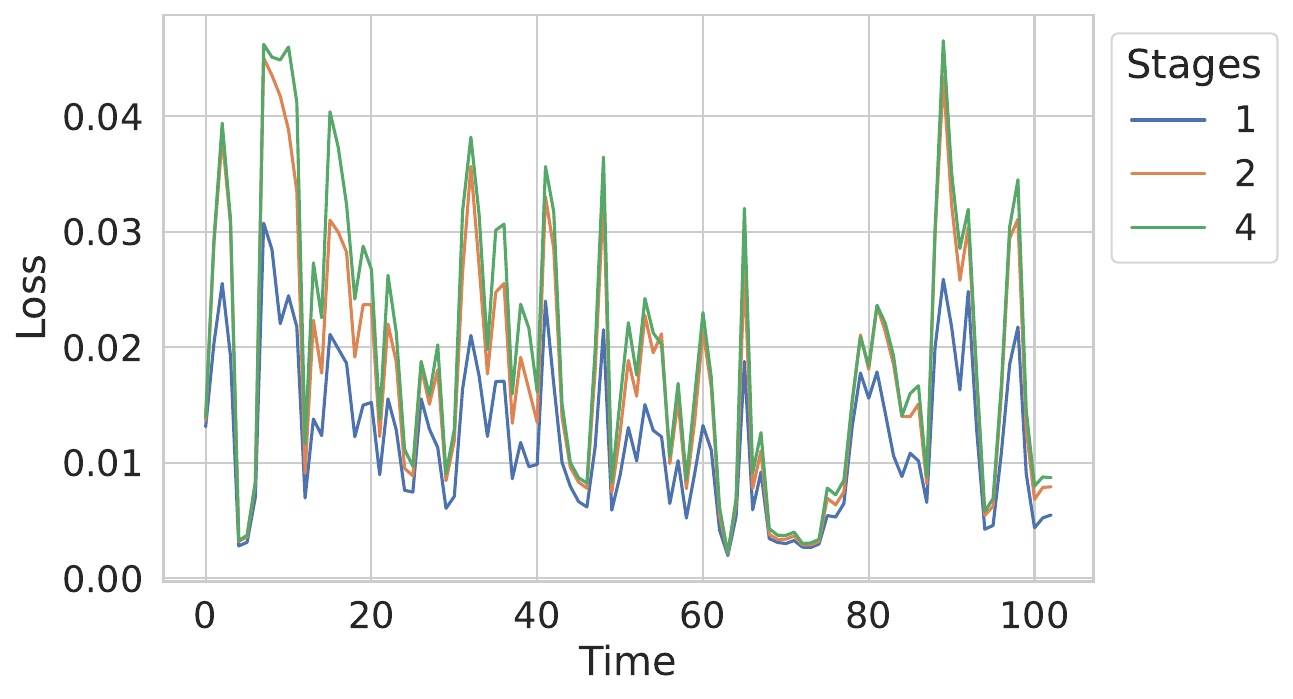}
        \caption{Continual online Optical Flow training loss over time (y-axes) over 1-minute windows on which the average loss is plotted (x-axis).}
        \label{fig:ps_loss}
    \end{figure}

    \noindent\textbf{E. Continual Online Optical Flow Estimation.}
    Our last experimental activity considers the optical flow learning problem \cite{brox2004high}, which consists of estimating the displacement field for all the pixels in a given frame pair. Such problem has been extensively investigated with the aid of deep neural networks, and we replicated the experience of \cite{colof}. 
    A convolutional neural network takes as input a frame pair (frames are concatenated along the channel dimension) and it outputs the displacement components for all the pixels. 
    Training is performed using a single frame pair at every step, and the pairs are sequentially extracted from a video source without shuffling. Unsupervised learning is driven by the brightness constancy assumption coupled with spatial regularization (see \cite{colof} for details). As video source, we choose ``1917”, a 2019 British war film directed and produced by Sam Mendes. The film lasts approximately 103 minutes (without credits) and appears as a single long continuous take, without artificial cuts.
    We report in Fig.~\ref{fig:flow_estimation} the optical flow estimated by the sequential and pipelined network.
    We selected a learning rate $\mu=10^{-5}$ for both the sequential and the 2-stage pipeline, while it is set to $\mu=10^{-6}$ for the 4-stage pipeline to account for the assumption of slowly changing gradients (see Section \ref{sec:local}). Noticeably, the estimated flow from the pipelined network is qualitatively similar to that of the sequential network, with little degradation in the 4-stages case. Fig.~\ref{fig:ps_loss} shows how the training loss changes every 1-minute window in the movie. Whilst the gradients computed by the pipelined model are more subject to approximation errors with an increased number of stages, we remark that the learning curve follows the same patterns of the vanilla sequential model, yielding a valid learning process.
    
    
    \newcommand{\framenumber}{85}
    \newcommand{\figwidth}{0.45}
    
    \begin{figure}
    
    \centering
        \subfloat[Frame\label{fig:flow_frame}]{%
            \includegraphics[width=\figwidth\linewidth]{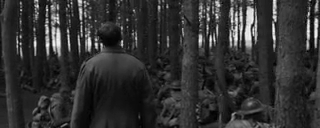}}
        \subfloat[Sequential\label{fig:flow_1stage}]{%
            \includegraphics[width=\figwidth\linewidth]{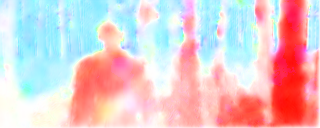}}\\
        \subfloat[2 Stages\label{fig:flow_2stage}]{%
            \includegraphics[width=\figwidth\linewidth]{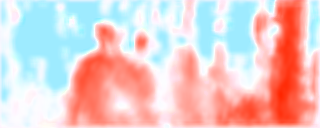}}
        \subfloat[4 Stages\label{fig:flow_4stage}]{%
            \includegraphics[width=\figwidth\linewidth]{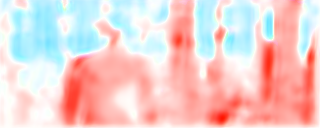}}\\
            
        \caption{Captures of Optical Flow estimation on a given frame (\ref{fig:flow_frame}) by the vanilla sequential network (\ref{fig:flow_1stage}), compared with the Optical Flow computed by the network handled by PARTIME  (\ref{fig:flow_2stage}, \ref{fig:flow_4stage}).}
    \label{fig:flow_estimation}
    \end{figure}
    
    \section{Conclusions and Future Work}
    \label{sec:conclusions}
    
    We presented and shared PARTIME, a Python software library designed for continual learning problems in which the data is streamed over time. PARTIME is built on a pipeline parallelism that speedups the computations of a neural network by a theoretical $D\times$, being $D$ the number of devices. We focused on the case of Graphics Processing Units (GPUs), showing that our implementation scales coherently with the expectations, as experimented in an environment with up to $8$ GPUs. Future work will consider and improve the implementation for specific classes of neural models, such as those that are recurrent and neural networks for graphs, as well to local optimization approaches \cite{lp}.

\bibliographystyle{IEEEtran}  
\bibliography{biblio}  

\end{document}